\documentclass[letterpaper, 10 pt, conference]{ieeeconf}  

\IEEEoverridecommandlockouts                              

\usepackage{graphicx} 
\graphicspath{{figures/}}
\usepackage{xcolor}
\usepackage{amsmath} 
\usepackage{amssymb}  
\usepackage{url}

\usepackage{amsthm}
\newtheoremstyle{remarkstyle}
  {\topsep}      
  {\topsep}      
  {\normalfont}  
  {}             
  {\bfseries}    
  {.}            
  { }            
  {}             
\theoremstyle{remarkstyle}

\title{\LARGE \bf
Composite-Gradient Learning for Shared Control Authority Between Deep Reinforcement Learning and Model Predictive Control
}

\author{Giray \"{O}n\"{u}r, Azita Dabiri and Bart De Schutter
\thanks{This research has received funding from the European Research Council (ERC) under the European Union's Horizon 2020 research and innovation programme (Grant agreement No. 101018826 - ERC Advanced Grant CLariNet).}
\thanks{The authors are with the Delft Center for Systems and Control,  Delft University of Technology, Delft, The Netherlands \{\tt\small g.oenuer, a.dabiri, b.deschutter\}@tudelft.nl.}%
}

\begin{document}

\maketitle
\thispagestyle{empty}
\pagestyle{empty}

\begin{abstract}
Integrated deep reinforcement learning (DRL) and model predictive control (MPC) methods are increasingly used to control autonomous systems by combining their complementary capabilities. DRL learns control policies through interaction with the environment. MPC uses a system model to optimize control inputs while accounting for constraints. In DRL-MPC frameworks with shared control authority, both the DRL agent and the MPC controller each determine part of the control inputs. However, common learning formulations treat MPC as part of the environment and therefore do not explicitly account for MPC's contribution to control or its interaction with the DRL agent. This paper proposes a novel composite-gradient learning (CGL) method that integrates the MPC controller into the learning process by representing the DRL and MPC control inputs as a joint action and accounting for their interaction when updating the DRL agent during training. CGL is evaluated on two multi-class freeway traffic networks with different strengths of interaction between the DRL and MPC control inputs and it is compared with alternative methods that treat MPC as part of the environment or that only partially incorporate MPC into learning. The results show that CGL offers limited benefit under weak interaction, but learns higher-performing control policies than the alternative methods in a subset of training runs under strong interaction, although the average control-performance gains remain modest.

\end{abstract}

\section{Introduction}

Model predictive control (MPC) and deep reinforcement learning (DRL) have become prominent approaches for controlling complex systems in applications ranging from industrial process control to robotics \cite{Rawlings2018,wang2022deep}. Although both methods address sequential decision making, they employ different approaches. MPC \cite{Rawlings2018} uses an explicit prediction model to forecast the system evolution over a finite horizon, optimizes a sequence of control inputs while explicitly accounting for system constraints, applies only the first control input of the optimized sequence, and repeats this procedure in a receding-horizon fashion. On the other hand, DRL \cite{wang2022deep} learns a control policy directly through interaction with the environment without requiring an explicit system model, allowing control policies to be obtained when model knowledge is limited or the system dynamics are uncertain. DRL also enables fast control input computation after training, since the trained policy network directly maps the current observation to an action through a single forward pass without requiring online optimization.

The complementary characteristics of DRL and MPC motivate hybrid frameworks that combine MPC's built-in optimization and constraint-handling capabilities with DRL's fast online computation and model independence \cite{reiter2026synthesis}. Hierarchical DRL-MPC frameworks exploit this combination by assigning different tasks and time scales to either method. Such frameworks improve control performance in automation applications such as traffic management \cite{sun2024novel} and microgrid energy management \cite{aljabri2026ami}. However, despite overall progress, integrating DRL and MPC remains a developing area with limited practical applications.

The current paper considers a hierarchical DRL-MPC framework \cite{onur2026dividing}, in which the control authority and control inputs are divided between DRL and MPC to allow MPC to run at a low frequency for control measures whose slower update rate accommodates MPC's high computation time, while DRL controls high-frequency control measures by leveraging its short deployment time. For example, high-frequency and low-frequency control inputs can correspond, respectively, to ramp metering rates and vehicle splitting rates in freeway networks \cite{pasquale2017multi}, actuator commands and reference trajectories in robotic motion control \cite{rosolia2022unified}, and semiconductor switching commands and motor-current references in power-electronic converter control \cite{xue2024multirate}. The hierarchical DRL-MPC framework therefore reduces online computation relative to purely MPC-based control while preserving MPC's optimization and constraint-handling  capabilities at the upper level. 

In this hierarchical DRL-MPC framework \cite{onur2026dividing}, the MPC controller is considered part of the environment while training the DRL agent, as treating the MPC controller as part of the environment is a common approach in such frameworks \cite{reiter2026synthesis}. However, when the control authority and control inputs are divided between MPC and DRL, since the MPC controller requires future DRL actions to predict the system evolution over its optimization horizon, the DRL agent influences the computation of the MPC control input. When MPC is treated as part of the environment, this dependence is not explicitly incorporated into the learned assessment of control performance or into the update of the DRL agent, thereby leaving the learning formulation incomplete. This may limit performance when the control inputs of the MPC controller and the DRL agent are strongly coupled.

For example, such coupling may arise in a freeway traffic control problem in which an MPC controller at the upper level uses route guidance to regulate the traffic flow directed toward a metered on-ramp, while a DRL agent at the lower level uses ramp metering to control vehicle entry onto the freeway mainline from that on-ramp. The high-level MPC control input determines the traffic demand arriving at the metered on-ramp and therefore the operating conditions faced by the lower-level DRL agent. Conversely, the predicted DRL actions over the MPC horizon influence the route guidance control input of the MPC controller by shaping the predicted traffic evolution. Together, these interactions create bidirectional coupling between the MPC controller and the DRL agent.

To incorporate this coupling into the learning process, we formulate the hierarchical control problem as a Markov decision process (MDP) in which the DRL action and the MPC control input form a joint action space. Based on this formulation, we propose a novel composite-gradient learning method that captures both the direct effect of the DRL action and its indirect effect through the MPC control input, thereby addressing the incomplete learning formulation that treats MPC as part of the environment \cite{reiter2026synthesis}.



The main contributions of this paper are as follows:
\begin{itemize}
\item We formulate a hierarchical DRL-MPC control framework with shared control authority as an MDP in which the lower-level DRL action and the upper-level MPC control input form a joint action to capture the interaction between the two control levels.

\item We propose a novel composite-gradient learning method in which the learned assessment of control performance accounts for the joint action, and the update of the DRL agent captures both the direct effect of the DRL action and its indirect effect through the MPC control input.

\item We develop a hierarchy-aware data storage and sampling scheme that reduces the computational cost of the learning process by reusing MPC computations across the data collected between two consecutive MPC control input updates.
\end{itemize}

The remainder of this paper is organized as follows. Section~\ref{sec:framework} describes the hierarchical DRL-MPC framework with shared control authority. Section~\ref{sec:composite-gradient} presents the proposed composite-gradient learning method. Section~\ref{sec:case-study} introduces the multi-class freeway traffic case studies and evaluates the proposed method under different DRL-MPC control input coupling conditions. Finally, Section~\ref{sec:conclusions} concludes the paper and outlines directions for future work.

\section{Hierarchical DRL-MPC Framework with Shared Control Authority}
\label{sec:framework}

\begin{figure}[!tbp]
\centering
\includegraphics[width=\columnwidth]{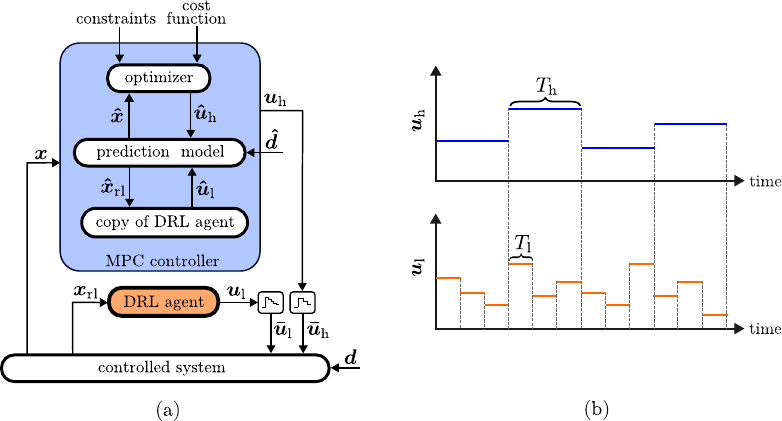}
\caption{(a) Block diagram illustrating the proposed DRL-MPC framework that divides control inputs between the DRL agent and the MPC controller. (b) Time scales for the low-frequency control inputs of the high-level MPC controller and the high-frequency control inputs of the low-level DRL agent. 
} \label{RL_MPC_simple}
\end{figure}

The hierarchical DRL-MPC framework divides control authority and control inputs between DRL and MPC \cite{onur2026dividing}, where the MPC controller uses a copy of the trained DRL agent, whose policy is held fixed during deployment, to obtain high-frequency control inputs required by MPC for predicting the future states of the controlled system (see Figure~\ref{RL_MPC_simple}(a)). This allows the MPC controller to predict the future states of the controlled system over its prediction window, even though the control input of the lower level is not determined by the MPC controller but by the DRL agent. Through this coupling, the DRL agent influences the MPC solution by affecting the state predictions used in the MPC optimization.

Let $\boldsymbol{x}(k)\in\mathcal{X}$ denote the state of the controlled system at sampling step $k$, where $\mathcal{X}$ is the state space of the system and $T$ is the length of the sampling time interval.
The high- and low-level control inputs jointly determine the evolution of the controlled system according to the discrete-time dynamical model
\begin{equation}\label{eq:system_dynamics}
\boldsymbol{x}(k+1)
= f\big(
\boldsymbol{x}(k),
\boldsymbol{\bar{u}}_\mathrm{h}(k),
\boldsymbol{\bar{u}}_\mathrm{l}(k),
\boldsymbol{d}(k)
\big),
\end{equation}
where $f$ denotes the system dynamics, $\boldsymbol{\bar{u}}_\mathrm{h}(k)$ and $\boldsymbol{\bar{u}}_\mathrm{l}(k)$ denote the high- and low-level control inputs applied to the controlled system at sampling step $k$, respectively, and $\boldsymbol{d}(k)$ denotes the exogenous disturbances.

Let $k_\mathrm{l}$ represent the control step counter for the low-level DRL agent, whose control update interval has length $T_\mathrm{l}$ (see Figure~\ref{RL_MPC_simple}(b)). The relationship between the low-level control interval and the sampling time interval is $T_\mathrm{l} = m_\mathrm{l} T$, $k = m_\mathrm{l} k_\mathrm{l}$, $m_\mathrm{l} \in \mathbb{N}^{+}$, $m_\mathrm{l} \geq 1$,
where $m_\mathrm{l}$ is the number of sampling intervals within one low-level control interval, $k=m_\mathrm{l}k_\mathrm{l}$ identifies the sampling step corresponding to low-level update step $k_\mathrm{l}$, and $\mathbb{N}^{+}$ denotes the set of positive integers. At every low-level control step, the DRL agent receives the observation $\boldsymbol{x}_\mathrm{rl}(k_\mathrm{l})\in\mathcal{X}_\mathrm{rl}$. Let $\boldsymbol{\eta}_\mathrm{rl}(k_\mathrm{l})$ denote auxiliary information available to the agent that is not necessarily contained in the current controlled-system state, such as exogenous measurements or previous states and control inputs. The DRL observation may combine selected or transformed components of the system state with this auxiliary information and is defined by the observation map $H_\mathrm{rl}$ as
\begin{equation}\label{eq:observation_map}
\boldsymbol{x}_\mathrm{rl}(k_\mathrm{l})
=H_\mathrm{rl}\big(
\boldsymbol{x}(m_\mathrm{l}k_\mathrm{l}),
\boldsymbol{\eta}_\mathrm{rl}(k_\mathrm{l})
\big).
\end{equation}

The DRL agent computes the high-frequency control inputs every $T_\mathrm{l}$ time units as $\boldsymbol{u}_\mathrm{l}(k_\mathrm{l}) = \pi_\theta\big(\boldsymbol{x}_\mathrm{rl}(k_\mathrm{l})\big)$, $\boldsymbol{u}_\mathrm{l}(k_\mathrm{l}) \in \mathcal{U}_\mathrm{l}$,
where $\pi_\theta$ is the deterministic DRL policy with parameters $\theta$, and $\mathcal{U}_\mathrm{l}$ is the admissible low-level control-input set. Then, the low-level control inputs that are applied to the controlled system during the control interval $[k_\mathrm{l}T_\mathrm{l},(k_\mathrm{l}+1)T_\mathrm{l})$ are calculated as $\boldsymbol{\bar{u}}_\mathrm{l}(m_\mathrm{l}k_\mathrm{l}+s) = \boldsymbol{u}_\mathrm{l}(k_\mathrm{l})$, $s \in \mathbb{I}_{m_\mathrm{l}}$,
where $\mathbb{I}_m = \{0,1,\dots,m-1\}$ for $m\in\mathbb{N}^{+}$. Hence, the control input $\boldsymbol{u}_\mathrm{l}(k_\mathrm{l})$ is applied over an interval of $T_\mathrm{l}$ time units using a zero-order hold (ZOH) strategy.

Let $\boldsymbol{u}_\mathrm{h}(k_\mathrm{h})$ denote the high-level control input applied at high-level control step $k_\mathrm{h}$ (see Figure~\ref{RL_MPC_simple}(b)), and let $\boldsymbol{u}_{\mathrm{h},j\mid k_\mathrm{h}}$ denote the high-level control input at high-level control step $k_\mathrm{h}+j$ calculated at high-level control step $k_\mathrm{h}$, where $j\in\mathbb{I}_{N_\mathrm{p}}$. The MPC controller computes the control sequence
\begin{equation*}
\boldsymbol{\tilde{u}}_\mathrm{h}(k_\mathrm{h}) = \left[ \boldsymbol{u}_{\mathrm{h},0\mid k_\mathrm{h}}^\top, \boldsymbol{u}_{\mathrm{h},1\mid k_\mathrm{h}}^\top, \dots, \boldsymbol{u}_{\mathrm{h},N_\mathrm{p}-1\mid k_\mathrm{h}}^\top \right]^\top
\end{equation*}
over a prediction horizon of \( N_\mathrm{p} \) high-level time steps, each with a duration of $T_\mathrm{h}$ time units, where $T_\mathrm{h} = m_\mathrm{h} T$, $k = m_\mathrm{h} k_\mathrm{h}$, $m_\mathrm{h} = q m_\mathrm{l}$, $q \in \mathbb{N}^{+}$.
Here, $q$ is the number of low-level control intervals within one high-level control interval, aligning the high-level and low-level control input updates (see Figure~\ref{RL_MPC_simple}(b)). At coincident high-level and low-level update time instants, their step counters therefore satisfy $k_\mathrm{l}=qk_\mathrm{h}$.

At high-level control time step $k_\mathrm{h}$, let
$\boldsymbol{\hat{x}}_{\ell\mid k_\mathrm{h}}
= \boldsymbol{\hat{x}}(m_\mathrm{h}k_\mathrm{h}+\ell)$ denote the
predicted future state of the controlled system $\ell$ sampling
steps ahead for
$\ell\in\mathbb{I}_{m_\mathrm{h}N_\mathrm{p}+1}$. The predicted future states are stacked as
\begin{equation*}
\boldsymbol{\tilde{x}}(k_\mathrm{h}) =
\left[
\boldsymbol{\hat{x}}_{1\mid k_\mathrm{h}}^\top,
\boldsymbol{\hat{x}}_{2\mid k_\mathrm{h}}^\top,
\dots,
\boldsymbol{\hat{x}}_{m_\mathrm{h}N_\mathrm{p}\mid k_\mathrm{h}}^\top
\right]^\top.
\end{equation*}
The states are predicted at the sampling steps to evaluate the
objective function accurately, even though the high-level control inputs
are updated only at the high-level control steps. The relative-index
notation is extended analogously to the predicted high-level inputs,
low-level inputs, and disturbances, denoted by
$\boldsymbol{\hat{u}}_{\mathrm{h},\ell\mid k_\mathrm{h}}$,
$\boldsymbol{\hat{u}}_{\mathrm{l},\ell\mid k_\mathrm{h}}$, and
$\boldsymbol{\hat{d}}_{\ell\mid k_\mathrm{h}}$, respectively, for
$\ell\in\mathbb{I}_{m_\mathrm{h}N_\mathrm{p}}$. The disturbance
forecasts $\boldsymbol{\hat{d}}_{\ell\mid k_\mathrm{h}}$
are assumed to be available to the MPC controller.
Let
$\boldsymbol{u}_{\mathrm{l},i\mid k_\mathrm{h}}$ and
$\boldsymbol{\hat{x}}_{\mathrm{rl},i\mid k_\mathrm{h}}$ denote the
predicted DRL action and the corresponding predicted DRL observation
$i$ low-level control steps ahead, respectively, where
$i\in\mathbb{I}_{qN_\mathrm{p}}$. Let
$\boldsymbol{\hat{\eta}}_{\mathrm{rl},i\mid k_\mathrm{h}}$ denote the
auxiliary information associated with that predicted DRL observation, which is obtained as $\boldsymbol{\hat{x}}_{\mathrm{rl},i\mid k_\mathrm{h}}
= H_\mathrm{rl}\big(
\boldsymbol{\hat{x}}_{m_\mathrm{l}i\mid k_\mathrm{h}},
\boldsymbol{\hat{\eta}}_{\mathrm{rl},i\mid k_\mathrm{h}}
\big)$, analogous to \eqref{eq:observation_map}.

The MPC controller computes the control inputs over the prediction horizon by solving the following optimization problem:
\begin{subequations}\label{eq:MPC}
\begin{align}
\min_{\boldsymbol{\tilde{u}}_\mathrm{h}(k_\mathrm{h}),\boldsymbol{\tilde{x}}(k_\mathrm{h})} \;\; 
& \sum_{\ell\in\mathbb{I}_{m_\mathrm{h}N_\mathrm{p}}}
L_\mathrm{mpc}\big(
\boldsymbol{\hat{x}}_{\ell\mid k_\mathrm{h}},
\boldsymbol{\hat{u}}_{\mathrm{h},\ell\mid k_\mathrm{h}},
\boldsymbol{\hat{u}}_{\mathrm{l},\ell\mid k_\mathrm{h}},
\boldsymbol{\hat{d}}_{\ell\mid k_\mathrm{h}}
\big) \nonumber \\
\text{s.t.}\;\;
& \boldsymbol{\hat{x}}_{\ell+1\mid k_\mathrm{h}} = \hat{f}\big(\boldsymbol{\hat{x}}_{\ell\mid k_\mathrm{h}}, \boldsymbol{\hat{u}}_{\mathrm{h},\ell\mid k_\mathrm{h}}, \boldsymbol{\hat{u}}_{\mathrm{l},\ell\mid k_\mathrm{h}}, \boldsymbol{\hat{d}}_{\ell\mid k_\mathrm{h}}\big), \nonumber\\
&\indent \text{for } \ell \in \mathbb{I}_{m_\mathrm{h}N_\mathrm{p}}, \nonumber \\
& \boldsymbol{\hat{x}}_{0\mid k_\mathrm{h}} = \boldsymbol{x}(m_\mathrm{h} k_\mathrm{h}), \label{eq:MPC_init} \\
& \boldsymbol{\hat{u}}_{\mathrm{h},m_\mathrm{h}j+s\mid k_\mathrm{h}} = \boldsymbol{u}_{\mathrm{h},j\mid k_\mathrm{h}}, \nonumber\\
&\indent \text{for } s \in \mathbb{I}_{m_\mathrm{h}}, \; j \in \mathbb{I}_{N_\mathrm{p}}, \label{eq:MPC_hold_h} \\
& \boldsymbol{\hat{u}}_{\mathrm{l},m_\mathrm{l}i+s\mid k_\mathrm{h}} = \pi_\theta \big(\boldsymbol{\hat{x}}_{\mathrm{rl},i\mid k_\mathrm{h}}\big), \nonumber\\
&\indent \text{for } s \in \mathbb{I}_{m_\mathrm{l}}, \; i \in \mathbb{I}_{qN_\mathrm{p}}, \label{eq:MPC_policy_l} \\
& \boldsymbol{\hat{x}}_{\ell\mid k_\mathrm{h}} \in \mathcal{C}_{\mathrm{x}}, \quad \ell \in \mathbb{I}_{m_\mathrm{h}N_\mathrm{p}+1} \backslash \{0\}, \label{eq:MPC_x_constraints} \\
& \boldsymbol{u}_{\mathrm{h},j\mid k_\mathrm{h}} \in \mathcal{C}_{\mathrm{u_h}}, \quad j \in \mathbb{I}_{N_\mathrm{p}}. \label{eq:MPC_u_constraints}
\end{align}
\end{subequations}
In \eqref{eq:MPC}, $L_\mathrm{mpc}$ denotes the stage-cost function, and $\hat{f}$ denotes the prediction model for the controlled system. Constraint \eqref{eq:MPC_init} sets the initial condition of the predicted state trajectory to the measured state at the current high-level control step. Constraint \eqref{eq:MPC_hold_h} maps the MPC control inputs from high-level control steps to sampling steps using a ZOH strategy. Constraint \eqref{eq:MPC_policy_l} specifies that the low-level control inputs used in the prediction are generated by a copy of the DRL policy based on the predicted DRL observation (see Figure~\ref{RL_MPC_simple}(a)) and held over the corresponding low-level control interval using a ZOH strategy. During deployment, the policy parameters $\theta$ remain fixed, and the same policy is used both to generate the low-level control inputs applied to the controlled system and to generate their predictions within the MPC problem. Finally, \eqref{eq:MPC_x_constraints} and \eqref{eq:MPC_u_constraints} collect the state and high-level input constraints, respectively. Here, $\mathcal{C}_{\mathrm{x}}\subseteq\mathcal{X}$ denotes the admissible state constraint set, and $\mathcal{C}_{\mathrm{u_h}}\subseteq\mathcal{U}_\mathrm{h}$ denotes the high-level control-input constraint set imposed by the MPC controller, where $\mathcal{U}_\mathrm{h}$ denotes the admissible high-level control-input set.

The resulting high-level control input corresponds to the first element of the optimized control sequence, $\boldsymbol{u}_\mathrm{h}(k_\mathrm{h})=\boldsymbol{u}^{\star}_{\mathrm{h},0\mid k_\mathrm{h}}$, and can be expressed as
\begin{equation}\label{eq:mpc_control_law}
\boldsymbol{u}_\mathrm{h}(k_\mathrm{h})
=
\pi_\mathrm{mpc}\left(
\boldsymbol{x}(m_\mathrm{h}k_\mathrm{h}),
\pi_\theta,
\boldsymbol{\eta}_\mathrm{mpc}(k_\mathrm{h})
\right),
\end{equation}
where $\pi_\mathrm{mpc}$ denotes the MPC control law obtained by solving \eqref{eq:MPC}, and $\boldsymbol{\eta}_\mathrm{mpc}(k_\mathrm{h})$ collects the time-varying auxiliary information required to solve the MPC problem that is not represented by the measured state or the DRL policy. Disturbance forecasts $\boldsymbol{\hat{d}}$ and predicted DRL-observation auxiliary information $\boldsymbol{\hat{\eta}}_{\mathrm{rl}}$ are examples of this type of information. 
The high-level control input is applied to the controlled system using a ZOH strategy as $\boldsymbol{\bar{u}}_\mathrm{h}(m_\mathrm{h}k_\mathrm{h}+s)=\boldsymbol{u}_\mathrm{h}(k_\mathrm{h})$, $s \in \mathbb{I}_{m_\mathrm{h}}$.

\section{Composite-Gradient Learning}
\label{sec:composite-gradient}

This section formulates the hierarchical control problem as an MDP and presents CGL together with a hierarchy-aware data storage and sampling scheme.

\subsection{MDP Formulation}
\label{sec:mdp-formulation}

Assuming Markov-sufficient observations, the learning problem is represented by the MDP tuple
\begin{equation*}
\left(\mathcal{X}_\mathrm{rl},\mathcal{U},\mathcal{P},r,\gamma\right),
\qquad
\mathcal{U}=\mathcal{U}_\mathrm{l}\times
\mathcal{U}_\mathrm{h}.
\end{equation*}
Here, $\mathcal{X}_\mathrm{rl}$ is the MDP state space formed by the DRL observations, $\mathcal{U}$ is the joint action space formed by the admissible low- and high-level control-input sets, $\mathcal{P}$ is the state-transition probability kernel, $r:\mathcal{X}_\mathrm{rl}\times\mathcal{U}\rightarrow\mathbb{R}$ is the one-step reward, and $\gamma$ is the discount factor. One MDP step corresponds to one low-level control interval.

The MDP state is the DRL observation
$\boldsymbol{x}_\mathrm{rl}(k_\mathrm{l})\in\mathcal{X}_\mathrm{rl}$.
The joint action $\boldsymbol{u}(k_\mathrm{l}) =[
\boldsymbol{u}^{\top}_\mathrm{l}(k_\mathrm{l}),\boldsymbol{u}^{\top}_{\mathrm{h}\mid\mathrm{l}}(k_\mathrm{l})
]^{\top} \in \mathcal{U}$ comprises the low- and high-level control inputs, both
represented on the low-level time scale. Unlike the DRL input, which is
recomputed at every low-level control step, the MPC input is updated once every
$q$ low-level control steps and held constant between updates. Accordingly, the
high-level input represented on the low-level time scale is denoted by
$\boldsymbol{u}_{\mathrm{h}\mid\mathrm{l}}(k_\mathrm{l})$ and defined as
$
\boldsymbol{u}_{\mathrm{h}\mid\mathrm{l}}(k_\mathrm{l})
=
\boldsymbol{u}_\mathrm{h}\left(
\left\lfloor k_\mathrm{l}/q \right\rfloor
\right),
$
where $\lfloor\cdot\rfloor$ denotes the floor operation, which returns the
greatest integer less than or equal to its argument.

The transition probability kernel $\mathcal{P}(\cdot\mid\boldsymbol{x}_\mathrm{rl},\boldsymbol{u})$ gives the conditional probability distribution of the next MDP state, represented by the next DRL observation. For a current DRL observation $\boldsymbol{x}_\mathrm{rl}$, a joint control input $\boldsymbol{u}$, and any measurable set $\mathcal{X}^{+}_\mathrm{rl}\subseteq\mathcal{X}_\mathrm{rl}$ of possible next observations $\boldsymbol{x}^{+}_\mathrm{rl}(k_\mathrm{l})=\boldsymbol{x}_\mathrm{rl}(k_\mathrm{l}+1)$, the transition probability kernel is given by
\begin{align*}
&\mathcal{P}\big(
\mathcal{X}^{+}_\mathrm{rl}\mid
\boldsymbol{x}_\mathrm{rl},\boldsymbol{u}
\big)\\
&=\Pr\!\left(
\boldsymbol{X}_\mathrm{rl}(k_\mathrm{l}+1)
\in\mathcal{X}^{+}_\mathrm{rl}
\ \middle|\
\boldsymbol{X}_\mathrm{rl}(k_\mathrm{l})
=\boldsymbol{x}_\mathrm{rl},
\boldsymbol{U}(k_\mathrm{l})=\boldsymbol{u}
\right),
\end{align*}
where $\Pr(\cdot\mid\cdot)$ denotes the conditional probability operator and $\boldsymbol{X}_\mathrm{rl}$ and $\boldsymbol{U}$ denote random variables with realized values $\boldsymbol{x}_\mathrm{rl}$ and $\boldsymbol{u}$, respectively. One transition covers the evolution of the controlled system and the auxiliary information from sampling step $m_\mathrm{l}k_\mathrm{l}$ to sampling step $m_\mathrm{l}(k_\mathrm{l}+1)$. Accordingly, $\mathcal{P}$ captures the controlled-system dynamics in \eqref{eq:system_dynamics} together with the evolution of the auxiliary information over this interval.

The reward $r$ evaluates the control performance associated with the MDP state and joint action over the $m_\mathrm{l}$ sampling intervals that constitute the low-level control interval.

\subsection{Learning Algorithm}

We adopt Deep Deterministic Policy Gradient (DDPG) \cite{lillicrap2015continuous} as the base actor-critic algorithm for developing composite-gradient learning (CGL), motivated by the successful application of DDPG across a range of control benchmarks \cite{wang2022deep} and, in particular, within the considered DRL-MPC framework \cite{onur2026dividing}. DDPG is suitable for the considered framework because its off-policy learning feature allows previously collected experience to be reused, reducing the repeated MPC computations required to collect new training experience.

Building on DDPG, CGL learns a deterministic DRL policy through an actor network $\pi_\theta$, parameterized by $\theta$, which maps the observed state to a DRL action. A critic network $Q_\phi$, parameterized by $\phi$, estimates the expected discounted return associated with a state-action pair under the current policy. To improve learning stability, CGL also maintains a target actor network $\pi_{\theta'}$ and a target critic network $Q_{\phi'}$, whose parameters $\theta'$ and $\phi'$ are updated gradually relative to those of the main networks. Past experiences are stored in a replay buffer $\mathcal{D}$, with the oldest experiences removed once the buffer reaches its capacity. At each network update, a mini-batch $\mathcal{M}$ comprising $M$ stored experiences is sampled from $\mathcal{D}$ and used to calculate the actor and critic updates.

For each low-level transition, CGL requires the MPC-initialization states associated with the current and next MDP states. The current MPC-initialization state is used to calculate the high-level control input and its contribution to the composite policy gradient (see \eqref{eq:current_policy_high_input} and \eqref{eq:high_input_actor_sensitivity}), whereas the next MPC-initialization state is used to construct the critic target (see \eqref{eq:critic_target} and \eqref{eq:target_mpc_control_input}).

\begin{samepage}
At low-level control step $k_\mathrm{l}$, the MPC-initialization state is defined as
\begin{equation}\label{eq:mpc_initialization_state}
\boldsymbol{x}_\mathrm{mpc}(k_\mathrm{l})
=
\boldsymbol{x}\left(
m_\mathrm{h}
\left\lfloor\frac{k_\mathrm{l}}{q}\right\rfloor
\right),
\end{equation}
\noindent which is the measured controlled-system state used as an input to $\pi_\mathrm{mpc}$ in \eqref{eq:mpc_control_law} at the most recent MPC update. Consequently, $\boldsymbol{x}_\mathrm{mpc}(k_\mathrm{l})$ remains unchanged over the $q$ low-level control steps between two consecutive MPC updates.
\end{samepage}

Let $k_{\mathrm{l},i}$ denote the low-level control step at which experience $i$ is collected. The MPC-initialization states associated with the current and next MDP states of experience $i$ are denoted by
\begin{equation*}
\boldsymbol{x}_{\mathrm{mpc},i}
=
\boldsymbol{x}_\mathrm{mpc}(k_{\mathrm{l},i}),
\qquad
\boldsymbol{x}^{+}_{\mathrm{mpc},i}
=
\boldsymbol{x}_\mathrm{mpc}(k_{\mathrm{l},i}+1),
\end{equation*}
respectively, and are identical unless the transition ends at a high-level control input update instant. CGL stores the augmented experience tuples
$(
\boldsymbol{x}_{\mathrm{mpc},i},\allowbreak
\boldsymbol{x}_{\mathrm{rl},i},\allowbreak
\boldsymbol{u}_{i},\allowbreak
r_i,\allowbreak
\boldsymbol{x}^{+}_{\mathrm{mpc},i},\allowbreak
\boldsymbol{x}^{+}_{\mathrm{rl},i},\allowbreak
\delta_i
)
$
in the replay buffer $\mathcal{D}$, where $\delta_i\in\{0,1\}$ indicates whether the transition is terminal.

For an experience $i$ sampled from the replay buffer, the critic target is calculated as
\begin{equation}\label{eq:critic_target}
y_i
=
r_i
+
\gamma(1-\delta_i)
Q_{\phi'}\left(
\boldsymbol{x}^{+}_{\mathrm{rl},i},
\boldsymbol{u}^{+}_{\theta',i}
\right),
\end{equation}
where
\begin{equation*}
\boldsymbol{u}^{+}_{\theta',i}
=
\begin{bmatrix}
\boldsymbol{u}^{+}_{\mathrm{l},\theta',i}\\
\boldsymbol{u}^{+}_{\mathrm{h}\mid\mathrm{l},\theta',i}
\end{bmatrix}
\end{equation*}
is the joint control input of the target policy associated with the next MDP state. The low-level control input of the target policy is calculated using the target actor network as
$
\boldsymbol{u}^{+}_{\mathrm{l},\theta',i}
=
\pi_{\theta'}(
\boldsymbol{x}^{+}_{\mathrm{rl},i}
).
$
The high-level control input of the target policy, represented on the low-level time scale, is calculated as
\begin{equation}\label{eq:target_mpc_control_input}
\boldsymbol{u}^{+}_{\mathrm{h}\mid\mathrm{l},\theta',i}
=
\pi_\mathrm{mpc}\left(
\boldsymbol{x}^{+}_{\mathrm{mpc},i},
\pi_{\theta'},
\boldsymbol{\eta}^{+}_{\mathrm{mpc},i}
\right),
\end{equation}
where $\boldsymbol{\eta}^{+}_{\mathrm{mpc},i}$ denotes the auxiliary information associated with the MPC update relevant to the next low-level control step. 

The loss for the critic network is calculated as
\begin{equation}\label{eq:critic_loss}
L(\phi)
=
\frac{1}{M}
\sum_{i=1}^{M}
\left(
y_i
-
Q_\phi\left(
\boldsymbol{x}_{\mathrm{rl},i},
\boldsymbol{u}_i
\right)
\right)^2,
\end{equation}
where $i$ indexes the $M$ experiences in the mini-batch $\mathcal{M}$. The critic parameters $\phi$ are updated by minimizing \eqref{eq:critic_loss} using a gradient-based optimizer, such as adaptive moment estimation (Adam) \cite{kingma2017adammethodstochasticoptimization}.

\begin{figure}[!t]
\centering
\includegraphics[width=\columnwidth]{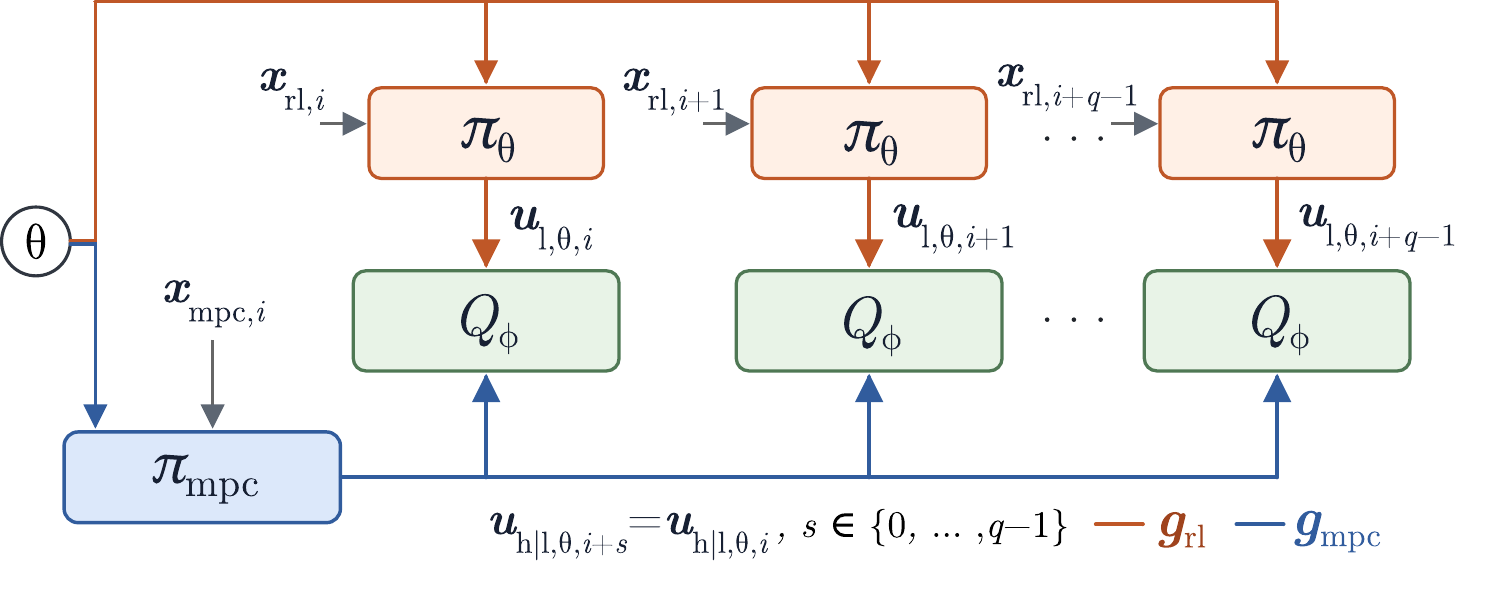}
\caption{Computational graph of the direct DRL contribution $\boldsymbol{g}_\mathrm{rl}$ and indirect MPC contribution $\boldsymbol{g}_\mathrm{mpc}$ within one group starting at $i$.}
\label{fig:computational-graph}
\end{figure}

The actor parameters $\theta$ are subsequently updated by maximizing the return estimated by the critic. For experience $i$, the joint control input induced by the current actor policy is defined as $\boldsymbol{u}_{\theta,i}
=
[\boldsymbol{u}^{\top}_{\mathrm{l},\theta,i},
\boldsymbol{u}^{\top}_{\mathrm{h}\mid\mathrm{l},\theta,i}]^{\top}
$. The low-level component is calculated as
$\boldsymbol{u}_{\mathrm{l},\theta,i}
=
\pi_\theta(\boldsymbol{x}_{\mathrm{rl},i})$.
The corresponding high-level component is calculated as
\begin{equation}\label{eq:current_policy_high_input}
\boldsymbol{u}_{\mathrm{h}\mid\mathrm{l},\theta,i}
=
\pi_\mathrm{mpc}\left(
\boldsymbol{x}_{\mathrm{mpc},i},
\pi_\theta,
\boldsymbol{\eta}_{\mathrm{mpc},i}
\right).
\end{equation} Unlike the joint input $\boldsymbol{u}_i$ stored in the replay buffer, $\boldsymbol{u}_{\theta,i}$ is calculated using the current actor parameters. The actor objective is defined as
\begin{equation}\label{eq:actor_objective}
J(\theta)
=
\frac{1}{M}
\sum_{i=1}^{M}
Q_\phi\left(
\boldsymbol{x}_{\mathrm{rl},i},
\boldsymbol{u}_{\theta,i}
\right).
\end{equation}
Based on \eqref{eq:actor_objective}, the composite policy gradient is decomposed as
\begin{equation*}
\nabla_\theta J(\theta)
=
\boldsymbol{g}_\mathrm{rl}
+
\boldsymbol{g}_\mathrm{mpc},
\end{equation*}
where $\boldsymbol{g}_\mathrm{rl}$ and $\boldsymbol{g}_\mathrm{mpc}$ denote the DRL and MPC contributions to the composite policy gradient, respectively (see Figure~\ref{fig:computational-graph}). The DRL contribution is calculated as
\begin{equation}
\boldsymbol{g}_\mathrm{rl}
=\frac{1}{M}\sum_{i=1}^{M}
\left(
\nabla_\theta
\boldsymbol{u}_{\mathrm{l},\theta,i}
\right)^{\!\top}
\nabla_{\boldsymbol{u}_\mathrm{l}}
Q_\phi\left(
\boldsymbol{x}_{\mathrm{rl},i},
\boldsymbol{u}_{\theta,i}
\right).
\label{eq:drl_gradient_contribution}
\end{equation}
The sensitivity $\nabla_\theta\boldsymbol{u}_{\mathrm{l},\theta,i}$ in \eqref{eq:drl_gradient_contribution} is obtained by differentiating the actor network with respect to the actor parameters $\theta$ as
$\nabla_\theta
\boldsymbol{u}_{\mathrm{l},\theta,i}
=
\nabla_\theta
\pi_\theta\left(
\boldsymbol{x}_{\mathrm{rl},i}
\right).$

The MPC contribution is calculated as
\begin{equation}
\boldsymbol{g}_\mathrm{mpc}
=\frac{1}{M}\sum_{i=1}^{M}
\left(
\nabla_\theta
\boldsymbol{u}_{\mathrm{h}\mid\mathrm{l},\theta,i}
\right)^{\!\top}
\nabla_{\boldsymbol{u}_{\mathrm{h}\mid\mathrm{l}}}
Q_\phi\left(
\boldsymbol{x}_{\mathrm{rl},i},
\boldsymbol{u}_{\theta,i}
\right).
\label{eq:mpc_gradient_contribution}
\end{equation}
The sensitivity $\nabla_\theta\boldsymbol{u}_{\mathrm{h}\mid\mathrm{l},\theta,i}$ in \eqref{eq:mpc_gradient_contribution} is obtained by differentiating the MPC control law\footnote{The MPC problem is assumed to admit a locally unique solution and to have a control law that is differentiable with respect to $\theta$ at the sampled experiences.} as
\begin{equation}\label{eq:high_input_actor_sensitivity}
\nabla_\theta
\boldsymbol{u}_{\mathrm{h}\mid\mathrm{l},\theta,i}
=
\nabla_\theta
\pi_\mathrm{mpc}\left(
\boldsymbol{x}_{\mathrm{mpc},i},
\pi_\theta,
\boldsymbol{\eta}_{\mathrm{mpc},i}
\right).
\end{equation}
Here, $\nabla_\theta\pi_\mathrm{mpc}$ denotes the derivative of the MPC control input with respect to the actor parameters through the policy $\pi_\theta$ embedded in the MPC problem. It therefore accounts for the effect of the actor parameters on the predicted low-level control inputs in \eqref{eq:MPC_policy_l} and, consequently, on the optimized high-level control input. Thus, the DRL contribution in \eqref{eq:drl_gradient_contribution} captures the direct effect of the DRL action on the estimated return, whereas the MPC contribution in \eqref{eq:mpc_gradient_contribution} captures the indirect effect of the DRL policy through the MPC control input, as illustrated in Figure~\ref{fig:computational-graph}. The actor parameters are then updated using gradient ascent as
\begin{equation}\label{eq:actor_parameter_update}
\theta
\leftarrow
\theta
+
\alpha_\theta\nabla_\theta J(\theta),
\end{equation}
where $\alpha_\theta>0$ denotes the actor learning rate.

Following the critic and actor updates, the target-network parameters are updated using Polyak averaging as in \cite{lillicrap2015continuous} to improve learning stability as
\begin{equation*}
\theta'
\leftarrow
\tau\theta+(1-\tau)\theta', \quad
\phi'
\leftarrow
\tau\phi+(1-\tau)\phi',
\end{equation*}

where $\tau\in(0,1)$ is the smoothing factor.

During experience collection, noise is added to the low-level DRL actions to encourage exploration. Before being applied to the controlled system, each resulting exploratory low-level action is projected into the admissible control-input set $\mathcal{U}_\mathrm{l}$. 

\begin{figure*}[!t]
\centering
\includegraphics[width=\textwidth]{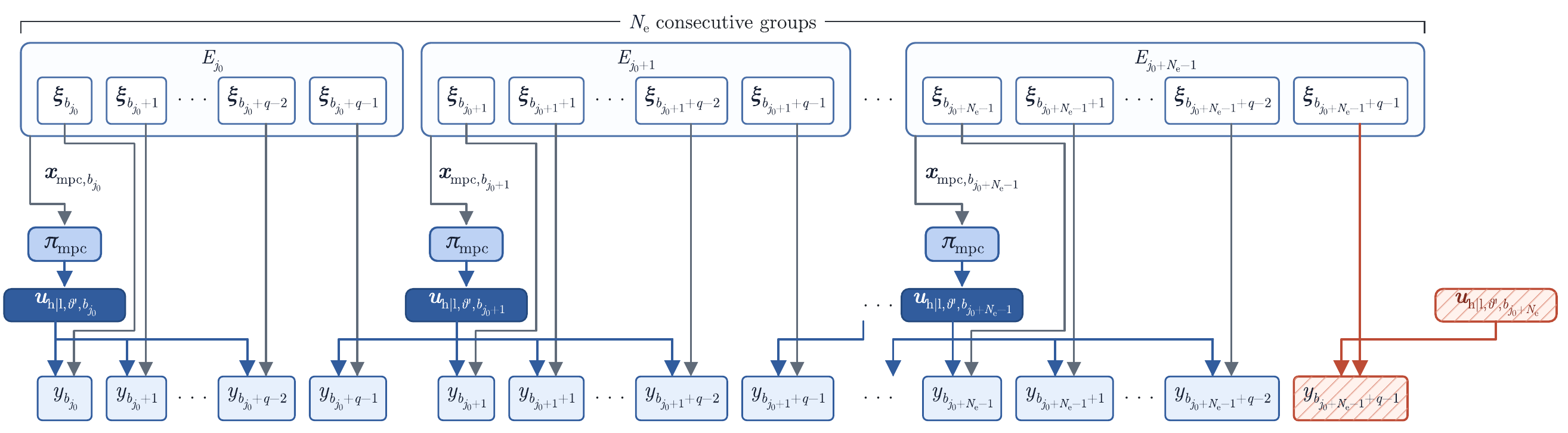}
\caption{Hierarchy-aware reuse of MPC evaluations under the target policy across consecutive experience groups. The red-hatched elements show the excluded final transition and its avoided additional MPC evaluation.}
\label{fig:hierarchy-target-flow}
\end{figure*}

\subsection{Hierarchy-Aware Data Storage and Sampling Scheme}
\label{sec:hierarchy-aware-sampling}

The critic target in \eqref{eq:critic_target} requires evaluating the MPC control law under the target actor policy according to \eqref{eq:target_mpc_control_input}. For the actor update in \eqref{eq:actor_parameter_update}, the MPC contribution in \eqref{eq:mpc_gradient_contribution} requires evaluating the control law under the current actor policy and calculating its sensitivity according to \eqref{eq:high_input_actor_sensitivity}. For fixed actor parameters, MPC evaluations with the same MPC-initialization state and auxiliary information yield identical control inputs and sensitivities. Accordingly, the proposed data storage and sampling scheme groups transitions by their associated MPC updates to reuse these computations during training.

Let $j$ index the groups, each containing the $q$ low-level transitions between two consecutive MPC updates. Let $b_j$ denote the index of the first experience in group $j$, so its experiences have indices $b_j+s$ for $s\in\mathbb{I}_q$ (see Figure~\ref{fig:hierarchy-target-flow}). The transition-specific entries of the augmented experience tuple are denoted by
$\boldsymbol{\xi}_i=(\boldsymbol{x}_{\mathrm{rl},i},
\boldsymbol{u}_i,
r_i,
\boldsymbol{x}^{+}_{\mathrm{rl},i},
\delta_i)$.
The replay buffer $\mathcal{D}$ stores these transitions in groups of the form
$E_j=(
\boldsymbol{x}_{\mathrm{mpc},b_j},
\{\boldsymbol{\xi}_{b_j+s}\}_{s\in\mathbb{I}_q}
)$.

Since the MPC-initialization state $\boldsymbol{x}_{\mathrm{mpc},b_j}$ is shared by all transitions in group $j$, it is stored only once. 
Experiences and groups are indexed in collection order, with $b_{j+1}=b_j+q$ for consecutive groups within an episode. According to \eqref{eq:mpc_initialization_state}, the next MPC-initialization state associated with transition $\boldsymbol{\xi}_{b_j+s}$ is given by
\begin{equation*}
\boldsymbol{x}^{+}_{\mathrm{mpc},b_j+s}
=
\begin{cases}
\boldsymbol{x}_{\mathrm{mpc},b_j},
& s<q-1,\\
\boldsymbol{x}_{\mathrm{mpc},b_{j+1}},
& s=q-1.
\end{cases}
\end{equation*}

For group $j$, the MPC control law in \eqref{eq:mpc_control_law} is evaluated under either actor policy as
\begin{equation*}
\boldsymbol{u}_{\mathrm{h}\mid\mathrm{l},\vartheta,b_j}
=
\pi_\mathrm{mpc}\left(
\boldsymbol{x}_{\mathrm{mpc},b_j},
\pi_\vartheta,
\boldsymbol{\eta}_{\mathrm{mpc},b_j}
\right),
\end{equation*}
where $\vartheta\in\{\theta,\theta'\}$ and $\boldsymbol{\eta}_{\mathrm{mpc},b_j}$ denotes the auxiliary information associated with the MPC update of group $j$.
The high-level control input of the current policy in \eqref{eq:current_policy_high_input} and its sensitivity in \eqref{eq:high_input_actor_sensitivity} are calculated once per group because their values are identical for all experiences in group $j$.

Using the MPC control inputs evaluated for each group, the high-level control input of the target policy in \eqref{eq:target_mpc_control_input} is obtained as
\begin{equation*}
\boldsymbol{u}^{+}_{\mathrm{h}\mid\mathrm{l},\theta',b_j+s}
=
\begin{cases}
\boldsymbol{u}_{\mathrm{h}\mid\mathrm{l},\theta',b_j},
& s<q-1,\\
\boldsymbol{u}_{\mathrm{h}\mid\mathrm{l},\theta',b_{j+1}},
& s=q-1.
\end{cases}
\end{equation*}
Consequently, the critic targets in \eqref{eq:critic_target} for the last transition of group $j-1$ and the first $q-1$ transitions of group $j$ use the same high-level control input of the target policy (see Figure~\ref{fig:hierarchy-target-flow}).

At each network update, $B$ sequences of $N_\mathrm{e}$ consecutive groups from the same episode are sampled uniformly from $\mathcal{D}$, with starting group indices $j_0^{(r)}$, $r\in\{1,\ldots,B\}$. The final transition of each selected sequence is excluded to avoid evaluating \eqref{eq:target_mpc_control_input} for the subsequent, unselected group, as illustrated in Figure~\ref{fig:hierarchy-target-flow}.
For tasks with a terminal condition, the final episode transition can instead be retained to learn that no future reward follows, using $y_i=r_i$.
The mini-batch is defined as
\begin{equation*}
\mathcal{M}
=
\biguplus_{r=1}^{B}
\left\{
\boldsymbol{\xi}_i
\mid
b_{j_0^{(r)}}\leq i<b_{j_0^{(r)}}+qN_\mathrm{e}-1
\right\},
\end{equation*}
where $\biguplus$ denotes multiset union, preserving repeated transitions if sampled sequences overlap. The $M=B(qN_\mathrm{e}-1)$ augmented experience tuples are reconstructed from the selected transitions and their group-level information, then relabeled with $i\in\{1,\ldots,M\}$ for the critic loss in \eqref{eq:critic_loss} and actor objective in \eqref{eq:actor_objective}. Each sequence requires $N_\mathrm{e}$ target-policy MPC evaluations for its $qN_\mathrm{e}-1$ critic targets in \eqref{eq:critic_target}. Likewise, $N_\mathrm{e}$ current-policy MPC evaluations and sensitivity calculations in \eqref{eq:high_input_actor_sensitivity} are reused across these transitions for the MPC gradient contribution in \eqref{eq:mpc_gradient_contribution}.

The number of MPC evaluations under the target policy per critic target using the proposed hierarchy-aware data storage and sampling scheme is therefore $N_\mathrm{e}/(qN_\mathrm{e}-1)$. If groups were sampled independently rather than consecutively, each group would require MPC evaluations under the target policy for both itself and its successor. Without reuse between sampled groups, retaining all $q$ transitions per group would therefore require $2N_\mathrm{e}$ evaluations for $qN_\mathrm{e}$ critic targets, giving $2N_\mathrm{e}/(qN_\mathrm{e})=2/q$ evaluations per target. For the proposed scheme, the ratio $N_\mathrm{e}/(qN_\mathrm{e}-1)$ decreases monotonically toward $1/q$ as $N_\mathrm{e}$ increases, approaching half the number of MPC evaluations per critic target required by independent sampling.
By reusing identical MPC computations under fixed network parameters, the proposed scheme reduces the MPC evaluation cost per retained transition without introducing additional approximations into its critic target or composite-gradient terms.

\section{Case Study}
\label{sec:case-study}

This section evaluates CGL on two multi-class freeway traffic networks and compares it with alternative methods.

\subsection{Freeway Traffic Control Settings}

\begin{figure}[!b]
\centering
\includegraphics[width=0.9\columnwidth]{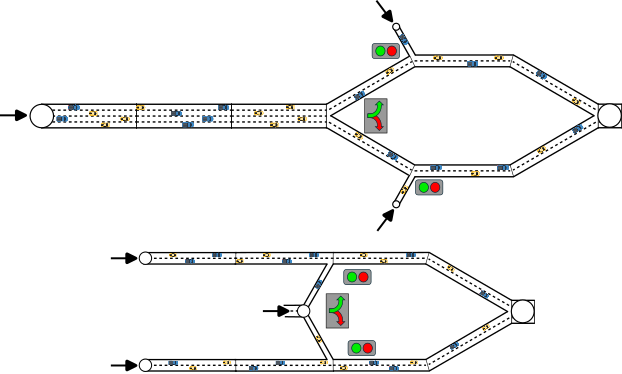}
\caption{Freeway layouts for Network~1 (top) and Network~2 (bottom). The green-and-red arrow symbol denotes the route split controlled by $u_\mathrm{h}$, and the traffic-light symbols denote the ramp-metering locations controlled by $u_{\mathrm{l},1}$ and $u_{\mathrm{l},2}$.}
\label{fig:case-study-networks}
\end{figure}

We consider the two freeway networks in Figure~\ref{fig:case-study-networks}, each with two vehicle classes. We use a multi-class METANET model \cite{pasquale2017multi} to simulate these networks with a sampling interval of $T=10$~s, owing to its balance between computational efficiency and modeling accuracy. Each 1-km link contributes seven state elements comprising the class-dependent mean speeds, densities, and outflows and the total density, while each origin contributes four state elements comprising the class-dependent queue lengths and outflows. Network~1 has nine links and three origins, yielding a 75-dimensional state vector $\boldsymbol{x}(k)$. Network~2 has eight links and four origins, yielding a 72-dimensional state vector. At each step, the demand vector $\boldsymbol{d}(k)$ contains $6$ class-specific demands of the three exogenous demand streams. 

Each network uses one vehicle-splitting input, $u_\mathrm{h}$, to distribute traffic between two routes and two ramp-metering inputs, $u_{\mathrm{l},1}$ and $u_{\mathrm{l},2}$, to regulate traffic entering the freeway from the two controlled on-ramps. The network structures yield different coupling strengths. In Network~1, $u_\mathrm{h}$ splits mainline traffic between two routes that later merge with separately metered on-ramps. In Network~2, $u_\mathrm{h}$ directly splits a shared demand stream between the two metered on-ramps, creating stronger bidirectional coupling with the ramp-metering actions.

The low-level DRL action
$\boldsymbol{u}_\mathrm{l}=[u_{\mathrm{l},1},u_{\mathrm{l},2}]^\top\in[0,1]^2$
determines two ramp-metering rates every $T_\mathrm{l}=120$~s. The high-level MPC input $u_\mathrm{h}\in[0,1]$ is a traffic splitting rate updated every $T_\mathrm{h}=600$~s. Hence, $q=5$, and each MPC input is held fixed while five successive DRL actions are applied. The MPC prediction horizon is 20~min, and each controlled episode lasts 30~min. Initial traffic conditions for both networks are established through a warm-up period before controlled operation.

\subsection{Learning Methods and Experimental Setup}

CGL is compared with three alternative learning methods: CGL-D (direct), CGL-S (scaled), and MPC-E. CGL-D uses the joint-action critic but updates the actor using only the direct gradient $\boldsymbol{g}_\mathrm{rl}$, omitting the indirect contribution $\boldsymbol{g}_\mathrm{mpc}$ through MPC. CGL-S uses $\boldsymbol{g}_\mathrm{rl}+\alpha\boldsymbol{g}_\mathrm{mpc}$ with $\alpha=\frac{n_\mathrm{h}\|\boldsymbol{g}_\mathrm{rl}\|}{n_\mathrm{l}\|\boldsymbol{g}_\mathrm{mpc}\|}$ to balance the DRL and MPC contribution norms in proportion to their control input counts, where $\|\cdot\|$ denotes the Euclidean norm and $n_\mathrm{l}$ and $n_\mathrm{h}$ are the dimensions of the low- and high-level control input vectors, respectively. MPC-E treats MPC as part of the environment, omits the high-level MPC input from the critic’s action input, and updates the actor using only $\boldsymbol{g}_\mathrm{rl}$.

The actor and critic use the reduced observation $\tilde{\boldsymbol{x}}_\mathrm{rl}$, which augments the traffic state with the current demand vector and the previously applied splitting rate. The reduced observation is treated as an approximate MDP state and has dimensions 82 and 79 for Networks~1 and~2, respectively.

The low-level reward over each 120-s interval is the negative sum of the total time spent (TTS), a queue-constraint penalty, and an input-change penalty. Here, the TTS is the aggregate time that vehicles spend traveling on the freeway links or waiting in the origin queues, so reducing it promotes efficient traffic movement. The queue-constraint penalty is one-sided and squared: no penalty is applied below a queue limit, and the penalty grows quadratically above it. Queue limits are 200 vehicles at mainline origins in both networks and 50 and 120 at each controlled on-ramp in Networks~1 and~2, respectively. Network~2 also includes a density-constraint penalty for the total density above $33$~veh/km/lane downstream of each controlled merge, with a weight of 2000. The input-change penalty applies to changes in all three control inputs with a weight of 0.4, encouraging smoother control operation. The reward is scaled by $1/30$ to control its magnitude, motivated by evidence that reward scaling can affect learning speed and asymptotic performance in off-policy actor-critic learning \cite{haarnoja2018soft}. MPC minimizes the negative of this reward over a prediction horizon of $N_\mathrm{p}=2$ high-level control steps, equivalent to 120 sampling steps, thereby aligning its objective with the DRL objective.

Similarly to \cite{sun2024novel}, the actor has two 256-unit hidden layers with rectified linear unit activations and layer normalization. The joint-action critic processes the observation and joint action through separate 256- and 128-unit branches, followed by 256- and 128-unit hidden layers. All methods use a replay-buffer capacity of 10\,000, a configured mini-batch size of 120, $\gamma=0.99$, actor and critic learning rates of $10^{-3}$, and $\tau=0.01$. The hierarchy-aware sampling scheme uses $N_\mathrm{e}=2$, yielding 108 retained transitions across $B=12$ sampled sequences per update. Decaying Ornstein-Uhlenbeck noise is added to the DRL actions during training, as suggested in \cite{lillicrap2015continuous}.

The nonlinear optimization problem in \eqref{eq:MPC} is solved using Optimistix \cite{rader2024optimistix} with at most 500 iterations. The sensitivity in \eqref{eq:high_input_actor_sensitivity} is computed by reverse-mode automatic differentiation of the finite-iteration solver map using a recursive checkpoint adjoint scheme, which, unlike implicit differentiation, does not require the solver to converge to a local minimum \cite{rader2024optimistix}.

Each method is trained using 10 independent neural-network-initialization and exploration seeds, with 960 complete episodes per run distributed over 64 parallel environments. The implementations and full evaluation outputs are available at \url{https://github.com/GirayOnur/composite-gradient-learning}.

\subsection{Results and Discussion}

\begin{figure}[!htbp]
\centering
\includegraphics[width=\columnwidth]{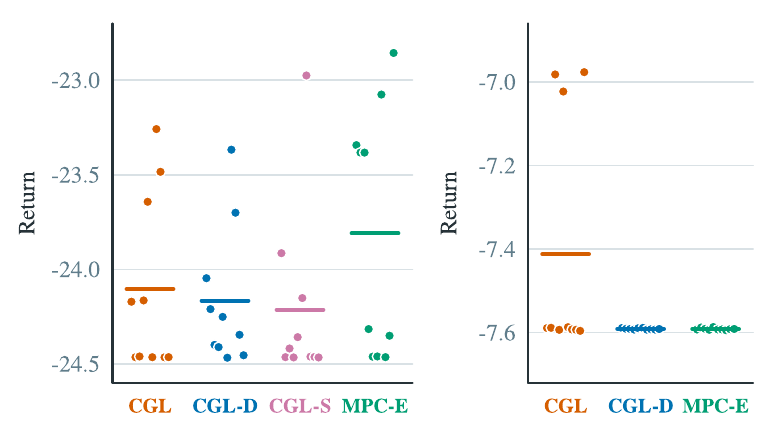}
\caption{Evaluation returns (higher is better) for Network~1 (left) and Network~2 (right). Each point shows the mean return of one trained policy over 100 evaluations with different demand-noise realizations. Horizontal bars show the mean over the 10 policies for each method.}
\label{fig:case-study-policy-returns}
\end{figure}

Figure~\ref{fig:case-study-policy-returns} compares the evaluation returns obtained by the learning methods in both networks. The return is the sum of the undiscounted rewards over one episode.

\begin{figure}[!htbp]
\centering
\includegraphics[width=\columnwidth]{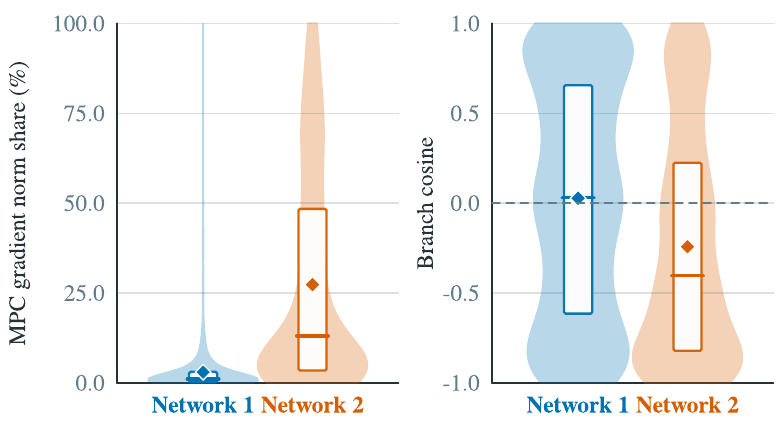}
\caption{Distributions of the norm share of the indirect MPC gradient (left) and branch cosine similarity between $\boldsymbol{g}_\mathrm{rl}$ and $\boldsymbol{g}_\mathrm{mpc}$ (right) during CGL training. Each network has 1\,720 logged updates (172 per policy); the Network~1 norm-share distribution excludes one zero-gradient update. Violin shapes, boxes, horizontal lines, and diamonds show the distributions, interquartile ranges, medians, and means, respectively.}
\label{fig:case-study-gradient-diagnostics}
\end{figure}

For Network~1, the return distributions overlap strongly, and the variation across trained policies is larger than the differences between the method means. Taking CGL as the reference, CGL-D has a lower mean return, whereas MPC-E has a higher mean return. The large overlap shows no clear advantage for any method in Network~1. For Network~2, the mean returns of CGL-D and MPC-E are both lower than that of CGL by approximately 2.4\% of its magnitude. Their trained policies reach nearly identical returns, while three of the 10 CGL policies achieve distinctly higher returns. These results suggest that CGL may learn higher-performing control policies than the alternative methods in a subset of training runs when the DRL and MPC control inputs interact strongly, although the average control-performance gains remain modest. The similar returns of CGL-D and MPC-E also show that including the high-level input in the critic is not sufficient to achieve this improvement; the actor update must also include the indirect contribution through MPC. Figure~\ref{fig:case-study-policy-returns} also shows that CGL-S does not improve on CGL in Network~1 despite amplifying $\boldsymbol{g}_\mathrm{mpc}$, whose magnitude is much smaller than that of $\boldsymbol{g}_\mathrm{rl}$ in this network (see also Figure~\ref{fig:case-study-gradient-diagnostics}). It is therefore not considered in Network~2, where the stronger DRL-MPC interaction results in a larger indirect MPC contribution.

Figure~\ref{fig:case-study-gradient-diagnostics} shows the norm share of the indirect MPC gradient, expressed as a percentage, and the cosine similarity for CGL during training. The norm share of the indirect MPC gradient is $\|\boldsymbol{g}_\mathrm{mpc}\|/(\|\boldsymbol{g}_\mathrm{rl}\|+\|\boldsymbol{g}_\mathrm{mpc}\|)$ expressed as a percentage, measuring the relative magnitude of the indirect MPC contribution. The cosine similarity measures the directional alignment between $\boldsymbol{g}_\mathrm{rl}$ and $\boldsymbol{g}_\mathrm{mpc}$, where positive values indicate similar directions and negative values indicate opposing directions.

In Network~1, the mean share of $\boldsymbol{g}_\mathrm{mpc}$ is only $3.0\%$, and the median cosine similarity between $\boldsymbol{g}_\mathrm{rl}$ and $\boldsymbol{g}_\mathrm{mpc}$ is 0.03. In Network~2, the mean share of $\boldsymbol{g}_\mathrm{mpc}$ increases to $27.3\%$, possibly due to the stronger interaction between the DRL and MPC control inputs, giving the indirect MPC contribution a larger role in learning. Moreover, the median cosine similarity decreases to $-0.40$, showing that $\boldsymbol{g}_\mathrm{mpc}$ often opposes $\boldsymbol{g}_\mathrm{rl}$, while including this opposing contribution can redirect the actor update toward the higher-return control policies shown in Figure~\ref{fig:case-study-policy-returns}.

\section{Conclusions}
\label{sec:conclusions}

This paper has proposed composite-gradient learning (CGL) for hierarchical control frameworks in which deep reinforcement learning (DRL) and model predictive control (MPC) share control authority. CGL integrates the MPC controller into the learning process by representing the DRL and MPC control inputs as a joint action and accounting for their interaction when updating the DRL agent during training. This paper has also proposed a hierarchy-aware data storage and sampling scheme that enables the reuse of MPC computations during training to reduce the computational burden of CGL.

In the freeway traffic control case studies, CGL offered limited benefit when the DRL and MPC control inputs interacted weakly. Under strong interaction, however, CGL learned policies that markedly outperformed the alternatives in some training runs, highlighting its potential to improve control performance despite modest average gains.

Future work will focus on detecting strong DRL-MPC control input coupling and selectively computing and using the composite gradient in such cases to improve the computational efficiency of CGL. 


\bibliographystyle{IEEEtran}
\bibliography{ref}

\end{document}